# Omnidirectional Three Module Robot Design and Simulation

Kartik Suryavanshi
Robotics Research Center
IIIT Hyderabad
Hyderabad, India
suryavanshikartik@gmail.com

Rama Vadapalli
Robotics Research Center
IIIT Hyderabad
Hyderabad, India
rama.raju@research.iiit.ac.in

Praharsha Budharaja
Robotics Research Center
IIIT Hyderabad
Hyderabad, India
Harsha.b@logicalsolutions.in

Abhishek Sarkar
Robotics Research Center
IIIT Hyderabad
Hyderabad, India
abhishek.sarkar@iiit.ac.in

Madhava Krishna
Robotics Research Center
IIIT Hyderabad
Hyderabad, India
mkrishna@iiit.ac.in

*Abstract—* **This paper introduces the Omnidirectional Tractable Three Module Robot for traversing inside complex pipe networks. The robot consists of three omnidirectional modules fixed 120° apart circumferentially which can rotate about their axis allowing holonomic motion of the robot. Holonomic motion enables the robot to overcome motion singularity when negotiating T-junctions and further allows the robot to arrive in a preferred orientation while taking turns inside a pipe. The singularity region while negotiating T-junctions is analyzed to formulate the geometry of the region. The design and motion capabilities are validated by conducting simulations in MSC ADAMS on a simplified lumped-model of the robot.**

## I. INTRODUCTION

Pipelines have a limited life cycle due to natural degradation or degradation due to static or dynamic loads. Early detection of damage in pipe networks can help in the prevention of catastrophic failures. Due to pipelines typically being concealed underground, it is difficult to inspect and determine the exact location of any damage. Numerous robots have been developed for pipeline inspection which can travel inside pipes and examine for any faults. One such embodiment is a fluid propelled robot called 'PIGs' [1], [2] (Pipe Inspection Gauges). These robots are pressure-driven and carry sensors for detection of pipe thickness or detection of any anomalies in the pipe. Since these robots are fluid-driven, the operator has no control over the direction, hence, these devices cannot be used in complex pipe networks.

Hirose, et al. [3] developed series of robots 'THES' for inspection of pipes with diameters ranging from Φ25mm to Φ150 mm and introduced 'Whole Stem Drive' with a series of modules to travel further in long winding thin pipes with multiple turns. Paulo, et al. [4] proposed PIPETRON, series of robots with articulated snake-like structure. PIPETRON I consisted of two wires, one which compressed the body used for clamping and other wire which provided yaw motion. In PIPETRON II the wires were replaced by torsion springs for compression and an additional actuator for yaw motion. Further, in PIPETRON VII, the yaw motion was made passive which made the robot come to the least energy state automatically while taking turns. Though the robot got simpler, the direction of motion could not be controlled in T junctions. Edwin, et al. [5] developed PIRATE with a similar configuration to target small diameter pipes and to turn in mitered bends. The robot design had two clamping V shapes links and omni wheels for holonomic motion along the principal axis. Kakogawa, et al. [6] developed a Screw Drive robot to pass through branch and bent pipes. The screw drive robots have a simpler structure and are easier to downsize. The robot used a differential mechanism for power transmission in between rolling and steering. However, the robot unintendedly steered in a straight pipe rather than rolling forward. Kakogawa, et al. [7], [8] later worked on the AIRO series of robots. The robots have articulated snake-like structure and omnidirectional hemispheres for holonomic motion. The rolling motion of the robot was ineffective when the hemispheres were in contact with the top surface. Also, the robot was unable to climb up in a joint of 60° joint angle. A flat in- pipe robot was developed by Young, et al. [9] with extra space for sensors mounting on the flat surface and a parallel linkage folding mechanism.

All the robots discussed above have driving modules fixed axially separated by a certain distance. Robots with three driving modules fixed circumferentially on extendable arms have shown superior mobility compared to other configurations [10]. MRINSPECT series has been the most successful robots with this configuration. MRINSPECT IV [10], a wheeled type robot, was developed with differential drive wheel for turning in curves. In MRINSPECT V [11], [12], a clutch system was developed for selective driving for greater efficiency during turns. Since the exact speeds of the individual driving modules in a turn are hard to calculate when turning, a 3-D mechanical differential was used in MRINSPECT VI [13]. Further, in MRINSPECT VI+ [14] rescue and brake mechanisms were added in the differential mechanism so that any slipping wheel does not consume all the power and the robot could be retrieved when stuck. A similar robot with a linkage-type mechanical clutch was developed by Young, et al. [15] for easy retrieval of the robot.

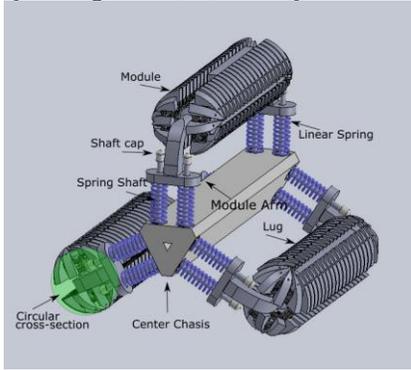

Fig.1: Omnidirectional Tractable Three Module Robot

Further, normal force control was explored in PAROYS-II [16], to increase the efficiency by actively changing the orientation in their robot. The unconstrained pantograph mechanism with active compliance joint made the track cover greater area. Atushi, et al. [17] also developed the underactuated parallelogram module robot which could climb through small steps inside pipes. Although the robots with modules fixed circumferentially show great mobility in a straight and curved pipe, they do not always have the provision to negotiate the T-junction [16], [18-20]. Young, et al. [18, 19] coined this inability as 'Motion Singularity'. In their work, the robot negotiates the T-junction by connecting another similar module oriented at 60° relative to the first module. Jong, et al. [20] discussed the motion singularity in the development of FAMPER and proposed two types of caterpillar mechanisms, one in which the caterpillar tracks are bent by 5° and the other in which caterpillar have a bendable structure. Our previous robot the Modular Pipe Climber has a similar configuration to the current design but lacked holonomic motion [21]. The Modular Pipe Climber successfully climbed vertically inside pipes and negotiated 45º and 90º bends with ease. However; the absence of holonomic motion meant that the Modular Pipe Climber could not orient itself to negotiate T-junctions [21].

We propose a robot with three omnidirectional modules, which is inspired by the work of Tadakuma, et al. [22], with the capability of holonomic motion in-plane about the robot central axis to overcome the motion singularity in T junction. The holonomic motion about the central axis enables the robot to avoid motion singularity in T-junction and also get to a preferred orientation while taking turns. Speed in the modules can be fixed to some standard values according to the geometry of standard turns. The description of the Omnidirectional Tractable Three Module Robot is presented in Section II. Section III proposes the singularity region and singularity sector is formulated. Section IV discusses the lumped model simulation and results

## II. DESCRIPTION OF THE ROBOT

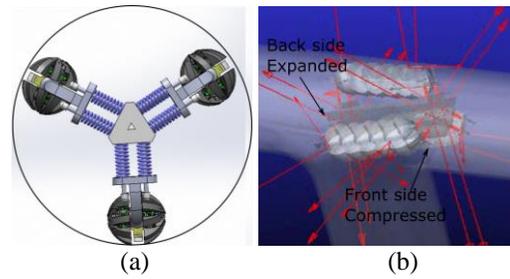

(a)          (b)

Fig.2(a) Circular cross-section of the Robot (b) Assymetrical compression

### A. Robot Structure

The robot is designed to climb vertically, turn inside bends and negotiate junctions of pipes with an inner diameter of 160mm. The robot has a circular cross-section of radius 90 mm at full extension of modules as shown in Fig.2. The robot structure consists of a triangular center chassis and three omnidirectional modules which are arranged 120° apart circumferentially. Each module slides on shafts attached to the center chassis and is pushed radially outward by the linear springs fixed outside the shafts. The module arms at the ends of the modules have slots bigger than the shafts (Fig.1) which provide the necessary tolerance for asymmetrical compression which helps in turning in pipe elbows as shown in Fig.2(b). The shaft caps restrict the module arms motion to a maximum specified length as shown in Fig.1.

### B. Module

The module consists of two motors one for driving the crawlers and other for rotation of modules about the central axis. Each module has two lug chain assemblies. The module components are described in the following subsections.

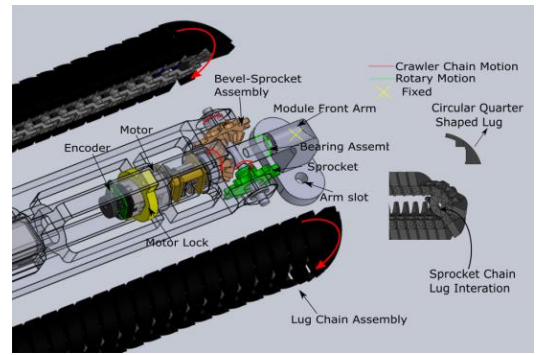

Fig.3 Exploded view of the front of the module

*a) Crawler Motor Assembly*

A single motor is used to drive two crawler chains (lug chain assembly) as shown in Fig.3. The motor is fixed along the central axis of the module and motion is transmitted by the bevel gear arrangement to drive both the crawler chains as shown in Fig.3. At the rear end (Fig.4), passively rotating sprockets are used to complete the crawler chain loops.

*b) Rotary Motor*

To rotate the module about their central axis (Fig.4), the rotary motor is fixed inside the module rather than fixing outside. It ensures that the module size remains compact. The motor shaft is locked with the rear arm and the motor body is fixed to the module chassis which rotates the module. When the motor shaft rotates in the clockwise direction, the motor body fixed to the module chassis rotates in the anticlockwise direction, hence the module also rotates in the anticlockwise direction.

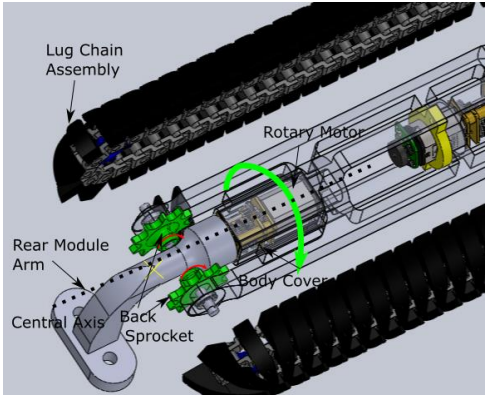

Fig.4 Exploded view of the rear of the module

*c) Crawler Chain Assembly*

A module has two crawler chains. Each crawler chain has a series of circular quarter-shaped studs called lugs (Fig.3). The design of the lug is such that the module forms a circular cross-section (Fig.1) for maximum contact with the wall of the circular pipe. The circular cross-section also allows for a smooth rolling of the module for the holonomic motion of the robot.

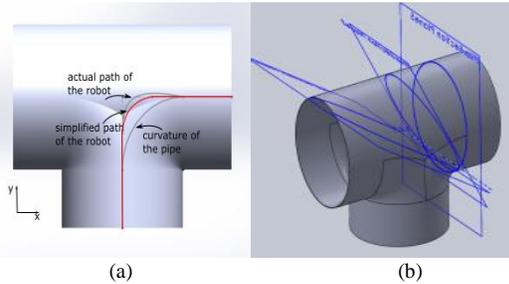

(a) (b)
Fig.5: (a) Robot actual and simplified path (b) Planes cutting the pipe.

## III. SINGULARITY REGION

During the vertical climb, all three modules compress equally such that the robot remains at the center of the pipe. But while negotiating the T junction, the actual path of the robot follows a curve similar to as shown in Fig.5(a) [23] which is above the curvature of the pipe. The actual path is difficult to determine as it is affected by the direction of gravity, turning direction and numerous other factors. For simplicity, we have assumed that the turning begins when the head reaches half of the radius of the pipe and motion of the robot is curvilinear following the simplified curve as shown in Fig.5(a), such that the robot covers equal distances in x and y directions in equal time steps as the robot rotates. The simplified path taken by the robot is the shown in Fig.5(a).

The planes perpendicular to the direction of motion of the robot at various instances are as shown in Fig.5(b). Fig.6 shows the cross-section of the pipes cut by the planes at 30°, 60° and 90° respectively. When an inclined plane intersects a cylindrical surface, an ellipse is formed. The major axis of the elliptical cross-section is greater than or equal to the pipe diameter as shown in Fig.6(a). As the robot traverses through the T junction, the eccentricity of the elliptical cross-section of the pipe goes on decreasing till it becomes zero, which is a circle, at the end of the turn. The changing cross-section of the pipe is shown in Fig.6(b).

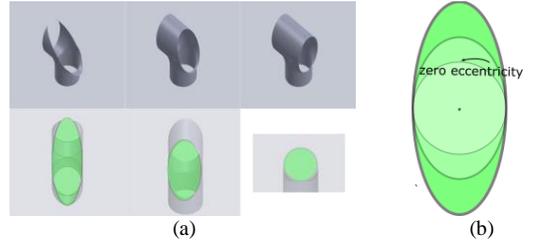

Fig.6: (a) Cross-section of the cut pipes (b) Changing eccentricity of the ellipse along the pipe

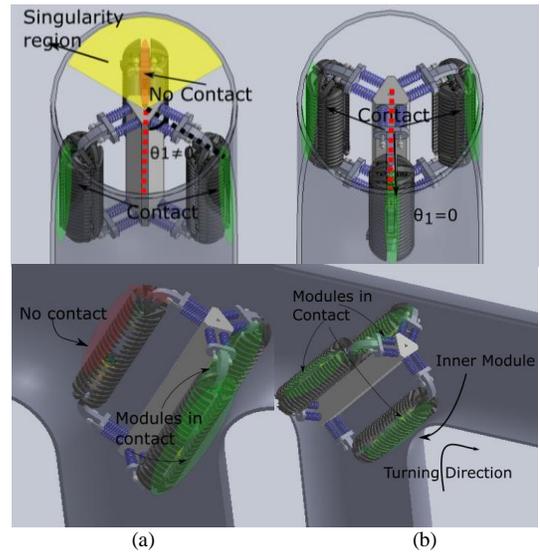

Fig.7: (a) Worst orientation (b) Best orientation

When the robot is negotiating the T junction, it has a inclined posture as shown in Fig.7. The orientation of the inclined robot is measured from the red dotted line in the circular cross-section of the pipe at the end of the curvilinear path. $\theta_1$ is the angle formed between the red dotted line and the inner module. We have considered the robot's preferred orientation as $\theta_1 = 0$ as shown in Fig.7(b). While negotiating T junction above a certain value of $\theta_1$, one of the outer modules even if the springs of the module are at free length (the module is fully extended) would not reach the pipe surface and maintain contact due to the elliptical cross-section of the pipe. The robot then would not have sufficient traction to negotiate the turn since only two modules are in contact with the pipe surface. This region is defined as the 'Singularity Region' while negotiating T junction which is shown in Fig.7(a) with yellow color. The region would form an elliptical sector on the elliptical cross-section. A projection of the elliptical sector is taken onto the last circular cross-section at the end of the curvilinear path to get a circular sector. The calculation of the 'Singularity Region' is shown below.

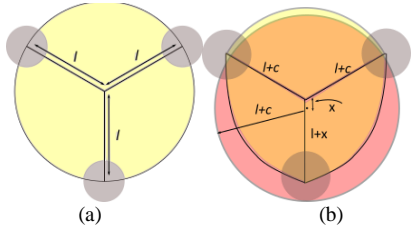

(a)          (b)

Fig 8: (a) The geometry in a Vertical climb (b)T junction

During vertical climb the cross-section of the robot is circular as all the springs compress equally Fig.8(a). But as the robot traverses in the T-junction with the preferred orientation (θ1 = 0), the inner module would carry most of the weight. Since the inner springs would compress more the outer modules would get space to expand and so the robot takes a deformed egg-shaped geometry as sown in Fig.8(b) in orange color.

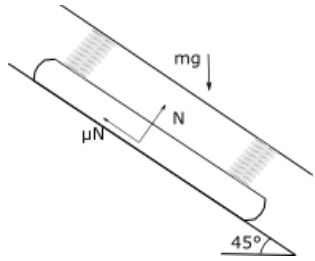

Fig.9: Forces on the inner module while negotiating T-junction

Assuming the inner module carries the weight of the robot, the four springs of the inner module would carry all the weight in effect during the turn. Equating forces in the vertical direction,

$$(\mu N + N)\cos 45 = W \quad (1)$$

where N is the normal force. The normal force compresses the four springs in the inner module,

$$4Ks = N \quad (2)$$

where the weight of the robot W = 7 N, the coefficient of the stiffness of spring K = 0.5 N/mm, the spring compression in the module (s) is calculated as 4 mm, which is less than the pre-compressed length, 5mm, in the springs to travel vertically. Therefore, the spring height in the inner module would increase by 1 mm relative to spring height during a vertical climb.

To formulate the 'Singularity Region', we find the contact points of the outer modules with the pipe surface at full extension. While traversing through the elliptical cross-section, the outer module would get more freedom to extend if the orientation of the robot is such that the outer module is closer to the 'Singularity Region', like in the Fig.7(a) where the outer module is completely inside singularity region and hence gets maximum space. Since the pipe surface becomes further, outer module would get more space to expand. As the lower module always maintains contact with the surface below, the geometry formed by the upper modules would be of interest and help us in determining the contact points with the pipe surface. For identifying the limits up to which the outer modules can reach without losing contact, we have taken the red circular region as the geometry of the robot as shown in Fig.8(b) which assumes the outer modules fully extended. Since the red circular region is formed by outer modules at fully extended state, the radius of the circle is equal to the radius of the robot when it is fully expanded.

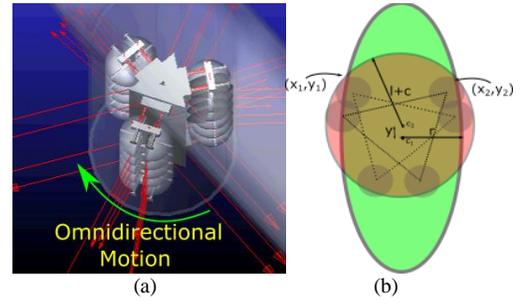

(a)          (b)

Fig.10: (a) In-plane motion (b) Possible limits of orientations.

When the robot is oriented such that the outer module just makes contact with the surface at free length, the robot would have one of the two orientations as shown in Fig.10 (b). Despite that springs of only one module would get space to expand to its maximum length, there would not be significant compression in the other outer module oriented away from the singularity region and we can assume that both the arms expand equally. With this assumption, the center of the circle would experience no horizontal shift and only vertical shift in the vertical direction by 1 mm as calculated by Equation 2.

The intersection points between the simplified robot circular geometry shifted up by 1 mm and the elliptical geometry of the pipe defines the limits of contact and hence the singularity region would lie between the limits $(x_1, y_1), (x_2, y_2)$ as shown in Fig.11. The equation of the ellipse formed by the intersection of the inclined plane and a cylinder is given by

$$x^2 + y^2 \cos^2 \theta = R^2 \quad (3)$$

where R is the radius of the cylinder (equal to radius of the pipe) and $\theta$ is the inclination angle of the plane. The equation of the circle shifted by 1 mm upwards in the y-direction is given by

$$x^2 + (y - 1)^2 = R'^2 \quad (4)$$

where R' is the radius of the red circular geometry.

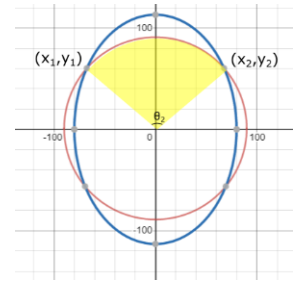

Fig.11 Singularity Sector

At the beginning of the curvilinear path, the cross-section ellipse would have an infinite eccentricity Fig.5(a) Although an ellipse with infinite eccentricity (θ = 90°, parallel lines) would have given us the sector with the largest area which would have been the worst case for motion singularity, all three modules would still be in contact with the straight section at the back of the robot.

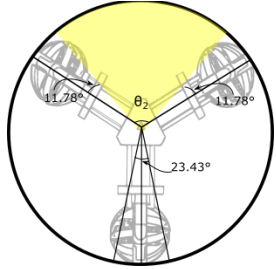

Fig.12 Available Region

Motion singularity is a condition when one of the modules loses contact which happens when the robot has left the straight section of the lower pipe and tilts towards the turn, encountering an elliptical cross-section. As it is difficult to determine the exact inclination angle when the robot leaves the bottom straight section and we know that by 45° the robot would have left the straight section at the bottom. Hence, we calculate the values at 45°. The intersection points plugging R = 80, R′ = 90 and θ = 45° are found as follows.

$$x_1 = -67.67, y_1 = 60.33 \text{ and } x_2 = 67.67, y_2 = 60.63$$

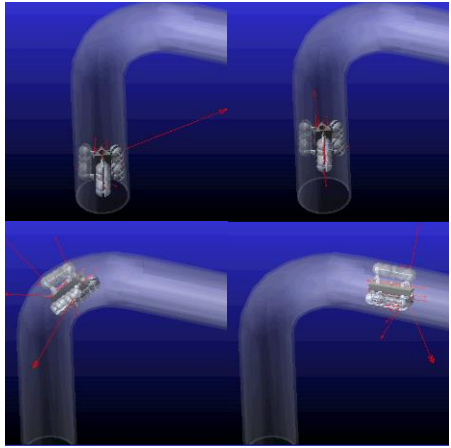

Fig.13: Robot in Elbow

The angle of the sector $θ_2$ is calculated from the center of the circle of robot's geometry as shown in Fig.11. The angle of the sector $θ_2$ is calculated as 96.564° which means the inner module has 23.43° angle to rotate and 11.71° in either direction from the preferred orientation (θ1 = 0) as shown in Fig.12.

## IV. IMPLEMENTATION AND EXPERIMENTATION

Simulations were conducted on Multi Body Dynamics software MSC Adams. A simplified lumped model was developed owing to the difficulty in handling the computational load with the actual complex model. In the lumped model the lugs were replaced by hemispherical balls to reduce the number of contact points which although could not capture the dynamics of the motion of the robot but was instrumental in kinematic and design feedback. The tests were conducted on pipes of a diameter of 160 mm. A constant speed of 100 mm/s was maintained in all the modules while traveling in straight pipes.

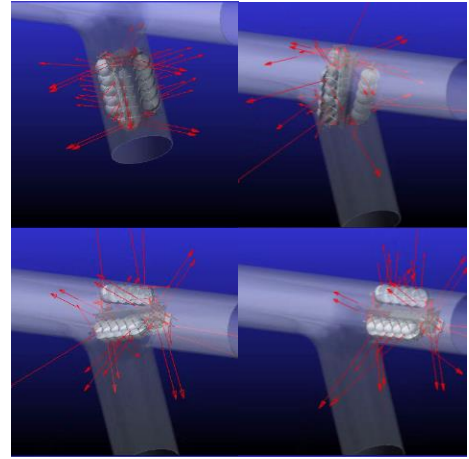

Fig.14: Simulation of the robot negotiating T junction

While turning in elbows (Fig.13), speed in the modules was set according to their traveling distance. The driving speeds were in the ratio of the respective radius of curvatures modules were taking. The driving ratios were 15:9:9 in our case. For negotiating the T junction (Fig.14), the robot is given a holonomic motion to get the robot to the preferred orientation which aligns one of the modules towards the turning direction and avoids the singularity region. Turning is initiated when the robot reaches half of the radius of the pipe. During the T junction, the inner module, which takes the shortest curvature is given velocity in the backward direction and the outer modules are driven forward. By driving the inner module in the backward direction, the robot is able to turn sharply. In Fig.16 we plot the values of angular velocities in the inner and the outer modules while negotiating T-junction. Opposite velocities during turning can be seen in the plot from about 0.7 s to 1.2 s. Although the speeds are modulated according to the geometry of the turn, slippage in the inner module is certain as finding the exact speed depends on points of contact, the number of lugs in contact, normal force distribution and other factors which makes it very difficult to calculate analytically.

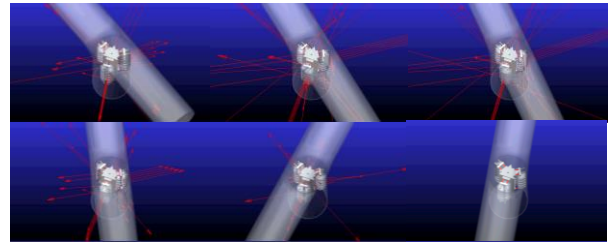

Fig.15: Holonomic motion

It was observed that when the modules are rotated about their central axis by 180° the crawler chain starts driving the robot in the opposite direction (Fig.17(a)). This happens because the crawling motion is opposite in the upper and lower side of the module, which explains the negative velocity input just before 0.5 seconds before holonomic maneuver and positive velocity input afterwards in Fig.16. To avoid this, the modules can be rotated in multiples of 360° or run in the opposite direction for every odd multiple of 180°. It was also observed that the robot does not crawl if the modules are at no motion line as shown in Fig.17(b).

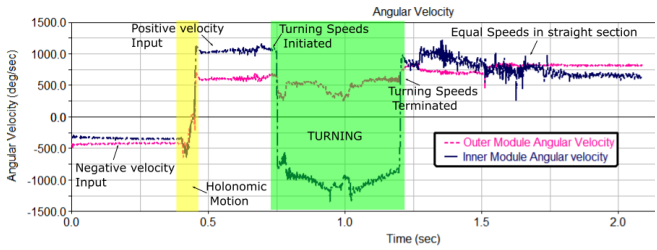

Fig.16: Angular velocity plot for outer and inner module

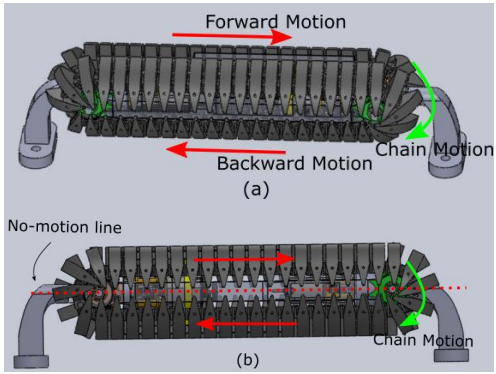

Fig.17: (a) Opposite motion (b) No-Motion Line

At this moment, since the contact is both with the upper and lower part of the chain loop, the module just wobbles in its location without any translation. This no-motion line is avoidable as it happens exactly at one angle.

## V. CONCLUSION AND FUTURE WORK

This paper introduces the Tractable Three Module Omni-crawler robot. The robot's motion capabilities allow the robot to negotiate T-junction and tackle the problem of 'motion singularity'. We identify the singularity region in the elliptical cross-section of the pipe. Kinematic simulations are conducted on a simplified model to verify the maneuvering capability of the robot. At present, we are working on making the physical embodiment of the robot and fabricating pipes for the same. We are also developing a 1-input 3-output differential for passive speed differential during motion in elbows.